\documentclass[10pt,twocolumn,letterpaper]{article}

\usepackage[pagenumbers]{cvpr} 

\usepackage{multirow}
\usepackage{wrapfig}

\definecolor{cvprblue}{rgb}{0.21,0.49,0.74}
\usepackage[pagebackref,breaklinks,colorlinks,allcolors=cvprblue]{hyperref}
\usepackage{booktabs}
\usepackage{colortbl}
\usepackage{colortbl}

\def\confName{CVPR}
\def\confYear{2027}

\title{MotionPatch: Motion-Aware Adversarial Patch Optimization in 4D Driving Scenes}

\author{\textbf{Jinlei Wang}$^{1,2,\dagger}$, \textbf{Jiahuan Long}$^{3,\dagger}$, \textbf{Jiacheng Hou}$^{2}$, \textbf{Yuanhao Huang}$^{4}$, \textbf{Mingkai Sun}$^{2}$, \textbf{Ming Wang}$^{2,5}$,\\
\textbf{Yang Yang}$^{2}$, \textbf{Wei Peng}$^{2}$, \textbf{Xu Liu}$^{2}$, \textbf{Chengyin Hu}$^{6}$, \textbf{Tingsong Jiang}$^{2,\ddagger}$, \textbf{Hongbo Chen}$^{1}$, \textbf{Wen Yao}$^{2,\ddagger}$\\
$^{1}$Sun Yat-sen University \quad $^{2}$Chinese Academy of Military Science \quad $^{3}$Shenzhen University\\
$^{4}$Beihang University \quad $^{5}$Shanghai Jiao Tong University \\
$^{6}$China University of Petroleum-Beijing at Karamay
}

\begin{document}
\maketitle
\begin{abstract}

Physical adversarial attacks against person detectors raise safety concerns for autonomous driving. Existing patch attacks often attach patches on 2D pedestrian images or optimize textures on 3D pedestrian models with fixed poses. However, these approaches struggle to model pedestrian motion and occlusion during optimization, which limit attack effectiveness in dynamic real-world environments. To address this, we propose MotionPatch, a motion-aware adversarial patch framework that incorporates pedestrian motion modeling into patch optimization to improve attack effectiveness in the physical world. Specifically, we reconstruct real driving scenes from RGB images and LiDAR point clouds using 3D Gaussian Splatting (3DGS). Based on these reconstructed scenes, we then insert animatable Gaussian pedestrians in these real driving scenes, with editable trajectories and poses to bridge the gap of digital-to-physical world. By utilizing scene depth information, we model the occlusion relationship between pedestrians and their surroundings. We also use 3D Geometry-Aware Illumination to adjust person appearance to match scene lighting. These strategies allow us to optimize adversarial patches under more realistic occlusion and lighting conditions. Experiments show that MotionPatch outperforms existing baselines and achieves SOTA performance against person detectors. Physical experiments further demonstrate its effectiveness under varying lighting, occlusion, and pedestrian motion conditions.

\end{abstract}
\section{Introduction}
\label{sec:intro}
Deep neural networks (DNNs) have improved image recognition~\citep{he2016deep, wang2024road} and natural language understanding~\citep{devlin2019bert}, but remain vulnerable to adversarial attacks.
Carefully crafted image perturbations can cause incorrect predictions even when humans can barely perceive the changes~\citep{goodfellow2014explaining,carlini2017towards, huang2024advswap, long2024papmot}.
In autonomous driving, this vulnerability poses a safety concern because visual perception is used to detect pedestrians and other road users~\citep{sun2020scalability}.

\begin{figure*}[t]
\centering
\includegraphics[width=\linewidth]{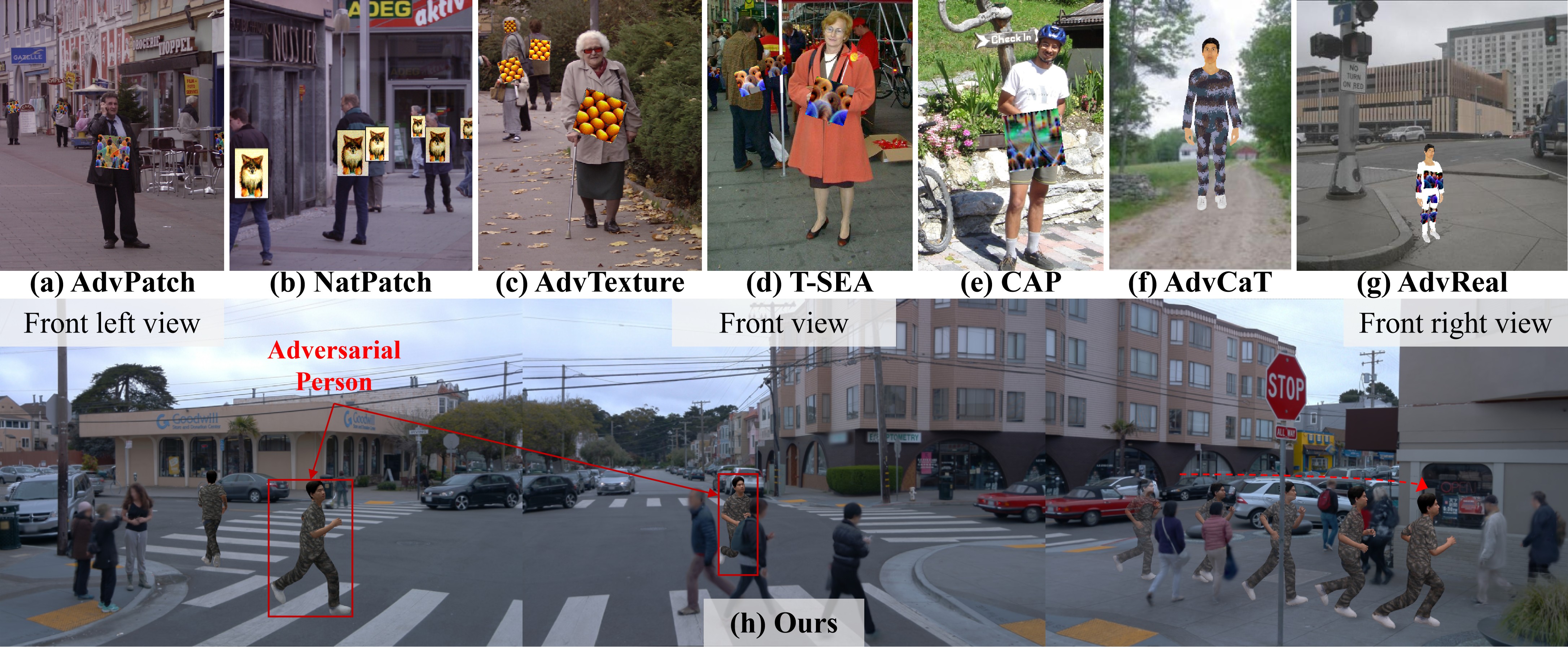}
\caption{Visual comparison of physical adversarial attacks on pedestrian detectors.
Panels (a--e) show patch-based methods, while panels (f--g) show methods based on 3D texture rendering. These methods may produce a visual mismatch between adversarial pedestrians and their surroundings. In contrast, our method (h) models pedestrian motion and occlusion in reconstructed 3DGS driving scenes, producing more realistic adversarial scenes.}
\label{fig:example}
\vspace{-5mm}
\end{figure*}

Recently, physical adversarial attacks have attracted increasing attention for evaluating the real-world robustness of person detectors. Existing attacks against person detectors can be broadly categorized into two types: image-based attacks and 3D Rendering-based adversarial attacks. Image-based methods optimize adversarial patterns by overlaying transformed patches onto pedestrian regions in 2D images~\citep{brown2017adversarial,thys2019fooling, hu2021naturalistic, huang2023t, wei2024revisiting, cheng2024full, wang2025transferable}. Thys et al. learn localized patterns that suppress person detections~\citep{thys2019fooling}, and Huang et al. improve attack transferability through self-ensemble optimization~\citep{huang2023t}. However, without explicit surface geometry, image-space transformations struggle to capture how adversarial patterns appear across substantial viewpoint and scale changes. 3D rendering-based methods map adversarial textures onto human or garment models and synthesize images from different camera viewpoints and distances. This enables geometry-aware optimization for attacks across multiple viewpoints and scales~\citep{hu2023physically, li2025uv, huang2026advreal, huang2026advserial, yan2026vfacamou}. Hu et al. optimize natural-looking clothing camouflage through 3D modeling~\citep{hu2023physically}, and Huang et al. vary the rendered pedestrian's placement, scale, viewpoint, and photometric appearance before compositing it with background images~\citep{huang2026advreal}. However, strong attack performance across viewpoints and scales does not necessarily ensure consistent effectiveness as pedestrians move through real-world scenes.

Despite this progress, existing 3D Rendering-based attack methods have three related limitations. First, pedestrian motion is not consistently modeled as a continuous trajectory coupled with scene geometry. Consequently, motion-induced changes in pose, viewpoint, and occlusion, and their joint effects on image formation, may be insufficiently represented. Second, rendered pedestrians are often composited with background images in image space rather than integrated with the scene geometry. This can produce a visual mismatch with the surroundings because depth-consistent occlusion is not enforced and simplified illumination may not reproduce local scene lighting. Third, optimizing aggregate attack performance can obscure failures on individual frames: a texture with strong average performance may still allow detections in some observations. Without explicitly coupling consecutive observations of the same pedestrian, consistent suppression over time is not guaranteed.

To address these limitations, we propose MotionPatch, which optimizes a shared adversarial garment texture in reconstructed 4D driving scenes. Specifically, we reconstruct real driving scenes from RGB images and LiDAR point clouds using 3D Gaussian Splatting (3DGS). Based on these reconstructed scenes, we then insert animatable Gaussian pedestrians in these real driving scenes, with editable trajectories and poses to bridge the gap of digital-to-physical world. By utilizing scene depth information, we model the occlusion relationship between pedestrians and their surroundings. We also use 3D Geometry-Aware Illumination to adjust person appearance to match scene lighting. These strategies allow us to optimize adversarial patches under more realistic occlusion and lighting conditions.

\begin{itemize}
    \item We propose MotionPatch, an adversarial patch framework that models continuous pedestrian motion in reconstructed 3DGS driving scenes. It supports inserting pedestrians with diverse appearances and editable trajectories and poses.

    \item We introduce 3D Geometry-Aware Illumination to match pedestrian appearance to scene lighting, improving visual consistency with the surrounding scene.
    
    \item We design a temporal continuity loss that couples consecutive observations of the same pedestrian and encourages consistent attack performance.

    \item Experiments show that MotionPatch outperforms existing person attack baselines. Physical experiments further demonstrate its effectiveness under varying lighting, occlusion, and pedestrian motion conditions.
\end{itemize}

\section{Related Work}

\begin{figure*}[t]
\centering
\includegraphics[width=0.85\linewidth]{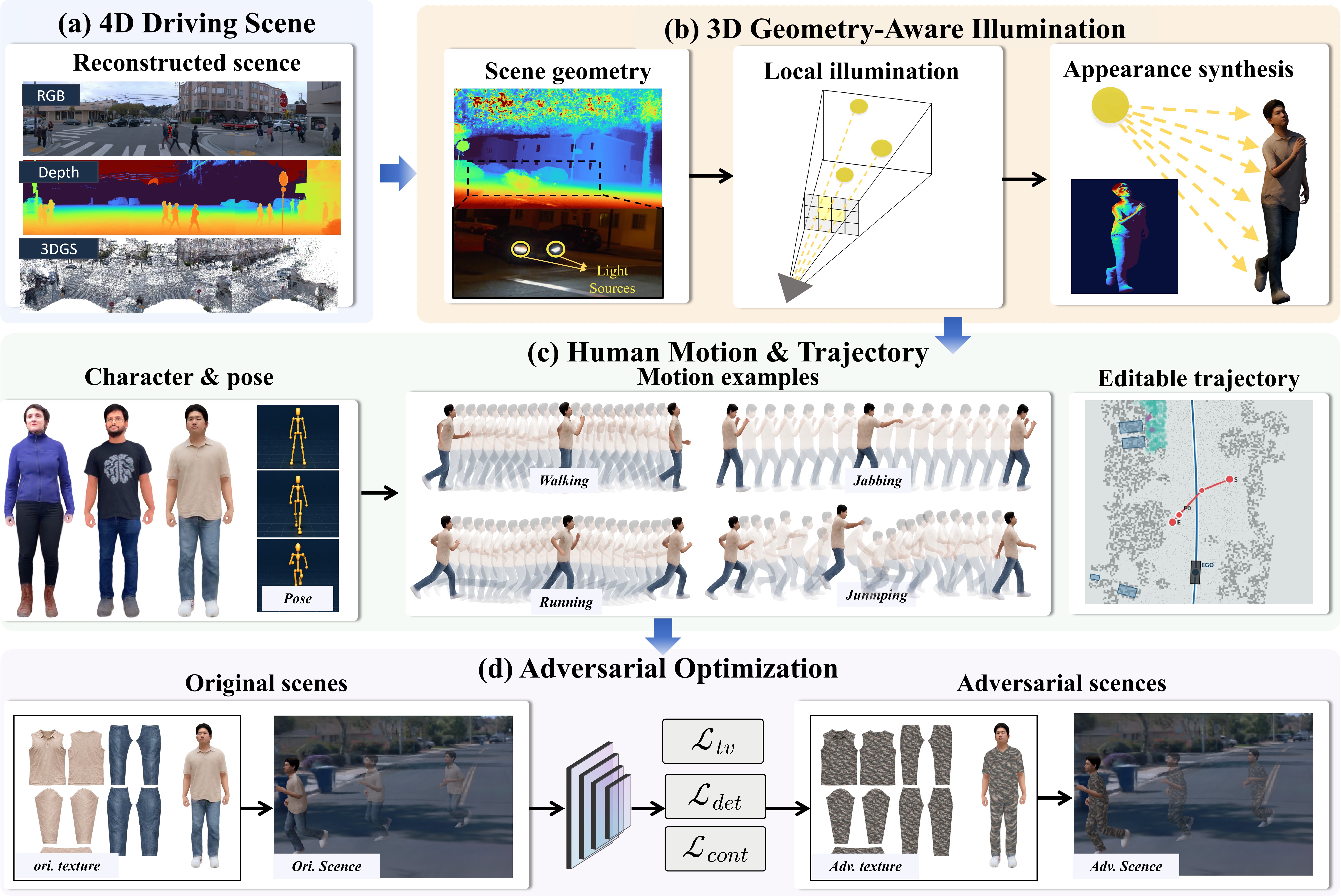}
\caption{Overview of the MotionPatch framework. (a) 4D Driving-Scene Reconstruction. Driving logs are reconstructed into 3D Gaussian Splatting scene graphs, providing RGB images and depth maps across timestamps and camera views. (b) 3D Geometry-Aware Illumination. Scene light backprojected into 3D space; their spatial relationships with pedestrian surface geometry determine local illumination for relighting. (c) Human Motion and Trajectory. Animatable avatar characters are driven by SMPL-X pose sequences sampled from a motion library and placed along manually annotated world-space trajectories within the reconstructed scenes. (d) Adversarial Optimization. Detection, temporal continuity, and total variation losses optimize an adversarial patch; the optimized patch is applied to avatars rendered within reconstructed 4D driving scenes.}
\label{fig-overview}
\vspace{-5mm}
\end{figure*}

\textbf{Image-Based Physical Adversarial Attacks.}~
Existing digital attacks~\citep{moosavi2016deepfool,chen2018ead,xie2019improving,li2020learning,hu2022adversarial, long2025robust,bai2025advglow} show that subtle input perturbations can cause deep neural networks to make incorrect predictions. Research has also extended to physical attacks, where adversarial patches alter an object's appearance before camera capture. Brown et al.~\citep{brown2017adversarial} introduced universal patches for physical scenes, and Thys et al.~\citep{thys2019fooling} extended this approach to person detection. Subsequent studies further improved adversarial patches in terms of visual naturalness, transferability, and physical robustness~\cite{long2024papmot, chen2025enhancing, long2025cdupatch, hu2025dynamicpae, long2026thermally}. Hu et al.~\cite{hu2021naturalistic} generated natural-looking patches using the latent space of a generative model. Huang et al.~\cite{huang2023t} introduced self-ensemble optimization to improve black-box transferability. Guesmi et al.~\cite{guesmi2024dap} employed dynamic patches and crease transformations to accommodate clothing deformation. More recently, Yan et al.~\cite{yan2026diff} leveraged class-optimized diffusion models to generate adversarial patches with improved visual naturalness and attack effectiveness. These image-based methods typically integrate transformed patterns with pedestrian regions through image-space compositing. Because the pattern is not explicitly attached to a 3D body or garment representation, this formulation does not model how its projection changes with pedestrian pose, viewpoint, and scale.

\textbf{3D Rendering-Based Physical Adversarial Attacks.}~
To improve physical attack robustness across poses and viewpoints, several methods bind adversarial textures to explicit 3D human or clothing mesh and render the resulting pedestrians under varied conditions. Hu et al.~\citep{hu2023physically} proposed AdvCaT, which parameterizes natural-looking camouflage with Voronoi patterns and learnable control points for rendering on a 3D human model. Li et al.~\citep{li2025uv} introduced UV-Attack, which uses dynamic-NeRF-based IUV maps to render editable clothing textures under sampled poses and views. Zhou et al.~\citep{zhou2025physically} modeled human--garment dynamics to generate sequences with physically simulated cloth deformation. Huang et al.~\citep{huang2026advreal} proposed AdvReal, which varies the rendered pedestrian's placement, scale, viewpoint, and photometric appearance before compositing it with a background image. Huang et al.~\citep{huang2026advserial} introduced AdvSerial, using skeletal animation and continuous-frame rendering for pedestrian attack sequences. Although these methods improve the representation of 3D-rendered appearance and, in some cases, pedestrian motion, the rendered foreground is still typically composited with a background image in image space. As a result, the rendered pedestrian's illumination may not adequately match the real scene, and occlusion is not necessarily resolved using the relative depth of the pedestrian and surrounding objects. Moreover, generating pose variations or continuous pedestrian sequences does not by itself account for changes in surrounding scene geometry and visibility over time.
\section{Methodology}
\label{sec:methodology}

MotionPatch optimizes a shared adversarial garment texture to evade pedestrian detectors in reconstructed 4D driving scenes. Figure~\ref{fig-overview} shows its four components: (a) 4D Driving Scene, (b) 3D Geometry-Aware Illumination, (c) Human Motion and Trajectory, and (d) Adversarial Optimization. We describe these components below.
\subsection{4D Driving-Scene Reconstruction}
\label{sec:scenario_construction}

We reconstruct each driving scene using OmniRe~\citep{chen2025omnire}, which represents the environment with static and dynamic 3D Gaussian components. Static components capture persistent scene structures, while dynamic components model content that changes over time. Although each timestamp is represented as a 3D scene, the sequence of time-indexed representations captures the scene's temporal evolution and thus forms a 4D driving-scene reconstruction, denoted by $\mathcal{G}_s(t)$.

For scene $s$, timestamp $t$, and camera $c$, we render an RGB background and its corresponding depth map:
\begin{equation}
(B_{s,t,c},D_{s,t,c})
=
\mathcal{R}_{\mathrm{GS}}
\left(\mathcal{G}_s(t);K_{s,t,c},T_{s,t,c}\right),
\label{eq:scene_rendering}
\end{equation}
where $K_{s,t,c}$ and $T_{s,t,c}$ denote the camera intrinsics and camera-to-world transformation, respectively. The RGB background provides the scene appearance used for compositing and the photometric cues for estimating illumination. The depth map supports two downstream operations. During pedestrian insertion, we compare the camera-space depth of each rasterized pedestrian surface point with the scene depth at the same pixel, retaining the nearer surface to resolve occlusion between the pedestrian and surrounding scene content.

\subsection{3D Geometry-Aware Illumination}
\label{sec:gai}

3D Geometry-Aware Illumination (GAI) relights the entire inserted pedestrian model using scene photometric statistics, surface geometry, and approximate local light sources. For each observation $(s,t,c)$, we render the posed pedestrian Gaussian avatar independently using the calibrated camera. Let $\Omega_P$ denote the pixels covered by this pedestrian-only render before scene compositing. At each $u\in\Omega_P$, the renderer provides the pedestrian color $A_P(u)$, its camera-space depth $D_P(u)$, and its world-space normal $\mathbf{n}(u)$. We backproject $D_P(u)$ using the same camera parameters to recover the world-space surface position $\mathbf{x}(u)$. The reconstructed-scene depth $D_{s,t,c}(u)$ is obtained from Eq.~\ref{eq:scene_rendering}; since both depth values refer to the same camera and pixel, they can be directly compared to determine visibility. The garment color is sampled from the optimized texture $P$, while other body regions retain their original colors.

\noindent\textbf{Scene-conditioned base shading.}~GAI estimates the global white level $\bar{w}$ from scene and a weak RGB tint $\boldsymbol{\tau}$ from bright, low-saturation pixels. A bounded gain $g$ accounts for local brightness variation across pedestrian placements. The resulting base shading combines ambient and normal-dependent directional terms:
\begin{equation}
\begin{aligned}
s_{\mathrm{geom}}(u)
&=
a_{\mathrm{amb}}
+
a_{\mathrm{dir}}
[\mathbf{n}(u)^\top\boldsymbol{\ell}_{\mathrm{dir}}]_+,
\\
E_{\mathrm{base}}(u)
&=
g\boldsymbol{\tau}
\min\!\left\{1,
\left(\frac{\bar{w}}{w_{\mathrm{ref}}}\right)^\gamma
\right\}\quad\cdot s_{\mathrm{geom}}(u).
\end{aligned}
\label{eq:gai_base}
\end{equation}
where $a_{\mathrm{amb}}$ and $a_{\mathrm{dir}}$ set the ambient and directional contributions, $\boldsymbol{\ell}_{\mathrm{dir}}$ is a prescribed unit direction, and $w_{\mathrm{ref}}$ and $\gamma$ control the brightness response.

\noindent\textbf{Geometry-aware local illumination.}~In dark scenes, GAI detects compact, locally contrasting bright regions with valid depth and backprojects them as approximate light sources. For source $j$ at world-space position $\boldsymbol{\ell}_j$, its response at the pedestrian surface is
\begin{equation}
\begin{aligned}
d_j(u)
&= \left\lVert \boldsymbol{\ell}_j-\mathbf{x}(u) \right\rVert_2,\\
\mathbf{v}_j(u)
&= \frac{\boldsymbol{\ell}_j-\mathbf{x}(u)}
{\max\!\left(d_j(u),\varepsilon\right)},\\
h_j(u)
&= \frac{
\left[\mathbf{n}(u)^\top\mathbf{v}_j(u)\right]_+
}{
1+\left(d_j(u)/r_j\right)^2
}.
\end{aligned}
\label{eq:gai_local_response}
\end{equation}
where $r_j$ controls spatial attenuation. GAI weights each response by the source RGB color, strength, and type-dependent gain. Low-height sources whose colors are not red-dominated also contribute camera-facing beams with angular weights. We clip the isotropic contributions before adding beam terms and bound the resulting local field $E_{\mathrm{loc}}(u)$; if no source is detected, $E_{\mathrm{loc}}(u)=0$.

\noindent\textbf{Appearance synthesis.}~Let $\Phi$ map sRGB to linear RGB, $\Gamma=\Phi^{-1}$, and $\rho_P(u)=\Phi(A_P(u))$. GAI applies the local term after base shading and clipping:
\begin{equation}
\begin{aligned}
\widetilde{\rho}_P(u)
&= \operatorname{clip}_{[0,1]}
\left(\rho_P(u)\odot E_{\mathrm{base}}(u)\right),\\
C_P(u)
&= \operatorname{clip}_{[0,1]}
\left\{\Gamma\left[
\widetilde{\rho}_P(u)
+\rho_P(u)\odot E_{\mathrm{loc}}(u)
\right]\right\}.
\end{aligned}
\label{eq:gai_color}
\end{equation}

Geometry and lighting estimates are held fixed during backpropagation, so gradients reach $P$ through texture sampling and the color operations. After relighting, we compare $D_P(u)$ with the corresponding scene depth $D_{s,t,c}(u)$: pedestrian contributions behind the reconstructed scene are suppressed, while visible pedestrian Gaussians are composited with the scene image $B_{s,t,c}$ according to their splatted opacity.

\subsection{Human Motion and Trajectory}
\label{sec:motion}

\noindent\textbf{Pedestrian representation.}~Each pedestrian is represented by an animatable human model whose geometry is encoded as Gaussian primitives bound to an SMPL-X mesh. Each Gaussian primitive $i$ is associated with a canonical appearance color $\mathbf{c}_{p,i}\in[0,1]^3$ and a canonical UV coordinate $\mathbf{u}_{p,i}\in[0,1)^2$. A UV-space mask $m_p(\mathbf{u})\in[0,1]$ identifies the garment regions assigned to the adversarial texture, including the torso, sleeves, cuffs, and trousers, while excluding the face, eyes, and uncovered hands. We use 14 such models reconstructed from single images: seven are pretrained models released with PERSONA~\citep{sim2025persona}, and the remaining seven are reconstructed from HuGe100K subjects using IDOL~\citep{zhuang2025idol}.

\noindent\textbf{Motion sequences.}~Pedestrian motion is represented by SMPL-X pose sequences derived from AMASS~\citep{mahmood2019amass}. Each sequence is retargeted to the pedestrian's body shape $\boldsymbol{\beta}_p$. For each pedestrian $p$ and motion clip, a frozen decoder $\mathcal{D}_p$ maps the SMPL-X pose $\boldsymbol{\theta}_{p,k}$ and body shape to a posed Gaussian representation:
\begin{equation}
\mathcal{A}_{p,k}=\mathcal{D}_p(\boldsymbol{\theta}_{p,k},\boldsymbol{\beta}_p).
\end{equation}
These posed representations are precomputed before optimization. At each training epoch, pedestrian identities are reassigned across scenes, and each pedestrian is assigned a motion clip sampled uniformly from those compatible with its trajectory type. A random temporal phase $\phi_p\in[0,1)$ varies the starting point of the assigned clip. For scene frame $t$, the corresponding motion-frame index is
\begin{equation}
k_p(t)=\left\lfloor (t-t_p^0)\,\rho+\phi_p N_p^{\mathrm{mot}}\right\rfloor \bmod N_p^{\mathrm{mot}},
\label{eq:motion_index}
\end{equation}
where $t_p^0$ is the pedestrian's first active scene frame, $N_p^{\mathrm{mot}}$ is the length of its assigned clip, and $\rho$ controls the motion playback rate relative to scene time. The posed Gaussians $\mathcal{A}_p(t)=\mathcal{A}_{p,k_p(t)}$ are retrieved by index at rendering time and provide surface normals for relighting.

\noindent\textbf{Editable trajectory.}~Each pedestrian is placed along a manually annotated world-space trajectory. The trajectory specifies the root position $\mathbf{r}_p(t)\in\mathbb{R}^3$ and yaw angle $\psi_p(t)$ at each scene frame. We place the posed Gaussians along this trajectory using the world-space position
\begin{equation}
\mathbf{x}_{p,i}(t)=s_p R_z(\psi_p(t))\left(\mathbf{x}_{p,i}^{\mathrm{pose}}(t)-f_p\mathbf{e}_z\right)+\mathbf{r}_p(t),
\label{eq:placement}
\end{equation}
where $\mathbf{x}_{p,i}^{\mathrm{pose}}(t)$ is the position of Gaussian primitive $i$ in $\mathcal{A}_p(t)$, $R_z(\psi_p(t))$ is the yaw rotation matrix, $s_p$ is the pedestrian scale, $f_p$ is a clip-specific floor offset, and $\mathbf{e}_z=(0,0,1)^\top$. The same rigid transform is applied to each Gaussian's rotation, and its scale is multiplied by $s_p$. Root positions follow the edited trajectory rather than the AMASS root translation. Trajectories are annotated using a bird's-eye-view editor and verified manually; details are provided in the supplementary material.

\subsection{Optimization Objectives}

\textbf{Detection loss.}~%
To evade person detection, we minimize the confidence of the person prediction whose bounding box has the highest Intersection over Union (IoU) with the ground truth box. 
Given the scene image $i$, the victim detector $\mathcal{D}$ outputs a set of bounding boxes $b_q^{(i)}$, each associated with a confidence score $\operatorname{Conf}_q^{(i)}$. The detection loss is defined as
\begin{equation}
\mathcal{L}_{\mathrm{det}}
=
\sum_{i}
\operatorname{Conf}_{q^*}^{(i)},
\quad
q^*
=
\operatorname*{arg\,max}_{q}
\operatorname{IoU}
\left(
gt^{(i)},
b_q^{(i)}
\right),
\label{eq:detection_loss}
\end{equation}
where $gt^{(i)}$ denotes the ground truth bounding box of the human in the scene $i$.

\noindent\textbf{Temporal continuity loss.}~To encourage consistent suppression across consecutive frames, the temporal continuity loss penalizes differences between adjacent detection scores. For a sequence of $T$ consecutive observations of the same pedestrian in the same scene and camera view, we define
\begin{equation}
\mathcal L_{\mathrm{cont}}
=
\frac{1}{T-1}\sum_{t=1}^{T-1}
\max\!\left(d_t^{\mathrm{h}}-d_t^{\mathrm{l}},\,0\right),
\label{eq:continuity_loss}
\end{equation}
where $d_t^{\mathrm{h}}$ and $d_t^{\mathrm{l}}$ denote the larger and smaller detection losses of adjacent observations, respectively. The lower score $d_t^{\mathrm{l}}$ is treated as a constant during backpropagation.

\noindent\textbf{Total variation loss.}~The total variation loss promotes smooth texture transitions:
\begin{equation}
\mathcal L_{\mathrm{tv}}
=
\operatorname{mean}\!\left(|\nabla_x P|\right)
+
\operatorname{mean}\!\left(|\nabla_y P|\right),
\label{eq:tv_loss}
\end{equation}
where $\nabla_x P$ and $\nabla_y P$ are horizontal and vertical finite differences, respectively, each averaged over all valid spatial and color-channel entries.

The overall objective combines detection, temporal continuity, and total variation losses:
\begin{equation}
\mathcal L_{\mathrm{total}}
=
\lambda_{\mathrm{det}}\mathcal L_{\mathrm{det}}
+
\lambda_{\mathrm{cont}}\mathcal L_{\mathrm{cont}}
+
\lambda_{\mathrm{tv}}\mathcal L_{\mathrm{tv}}.
\label{eq:total_loss}
\end{equation}
where $\lambda_{\mathrm{det}}$, $\lambda_{\mathrm{cont}}$, and $\lambda_{\mathrm{tv}}$ are the weights for the detection, temporal continuity, and total variation losses, respectively.


\section{Experiments}
\label{sec:experiments}

\definecolor{oursbg}{RGB}{253,231,231}
\begin{table*}[t]
\caption{White-box comparison (ASR \%) of representative adversarial attacks on the Waymo, nuScenes, and Argoverse 2 datasets with different detectors.}
\centering
\footnotesize
\setlength{\tabcolsep}{3pt}
\renewcommand{\arraystretch}{1.0}
\setlength{\arrayrulewidth}{0.6pt}
\begin{tabular}{l|c|c|ccc|ccc|ccc|c}
\toprule
\multirow{2}{*}{\textbf{Method}} & \multirow{2}{*}{\textbf{Venue}} & \multirow{2}{*}{\textbf{Year}}
 & \multicolumn{3}{c|}{\textbf{Waymo}} & \multicolumn{3}{c|}{\textbf{nuScenes}} & \multicolumn{3}{c|}{\textbf{Argoverse 2}} & \multirow{2}{*}{\textbf{Avg.}} \\
\cmidrule(lr){4-6}\cmidrule(lr){7-9}\cmidrule(lr){10-12}
 & & & YOLOv5 & F-RCNN & D-DETR & YOLOv5 & F-RCNN & D-DETR & YOLOv5 & F-RCNN & D-DETR & \\
\midrule
AdvPatch\cite{thys2019fooling}   & CVPRW         & 2019 & 40.74 & 25.72 & 36.36 & 15.45 & 4.43 & 17.61 & 24.95 & 15.71 & 30.08 & 23.45 \\
NatPatch\cite{hu2021naturalistic}   & ICCV          & 2021 & 24.52 & 17.45 & 31.25 & 9.77 & 8.99 & 15.51 & 18.53 & 15.10 & 33.24 & 19.37 \\
AdvTexture\cite{hu2022adversarial} & CVPR          & 2022 & 50.42 & 44.67 & 50.22 & 24.73 & \underline{22.21} & 31.44 & 38.94 & \textbf{37.72} & \underline{56.50} & 39.65 \\
T-SEA\cite{huang2023t}      & CVPR          & 2023 & 38.16 & 21.47 & 33.89 & 4.72 & 2.31 & 6.91 & 27.27 & 11.48 & 29.28 & 19.50 \\
DePatch\cite{cheng2024depatch}    & arXiv         & 2024 & 47.02 & 31.06 & 48.28 & 12.88 & 9.46 & 25.55 & 13.81 & 9.43 & 16.19 & 23.74 \\
CAP\cite{wei2024revisiting}        & NeurIPS       & 2024 & \underline{66.78} & 41.19 & 52.34 & 16.75 & 7.65 & 25.12 & 44.74 & 14.81 & 18.90 & 32.03 \\
DM\cite{lin2025diffusion}         & ICIP          & 2025 & 53.67 & 48.79 & 43.02 & 27.44 & 11.94 & 24.73 & \underline{46.07} & 28.89 & 55.94 & 37.83 \\
BadPatch\cite{wang2025badpatch}   & ICCVW         & 2025 & 51.95 & 36.51 & 51.78 & 27.95 & 8.65 & 28.50 & 15.64 & 10.84 & 21.38 & 28.13 \\
PosePatch\cite{zhou2025fooling}  & Neurocomputing & 2025 & 33.83 & 34.02 & 44.60 & 35.23 & 12.06 & \underline{38.78} & 12.56 & 15.94 & 17.49 & 27.17 \\
TRDPatch\cite{wang2025transferable}   & IOTJ          & 2025 & 46.00 & 38.75 & 43.17 & 17.54 & 13.54 & 25.46 & 21.60 & 15.69 & 30.45 & 28.02 \\
FABAttack\cite{hou2025fab}  & ACMMM         & 2025 & 63.34 & \underline{58.63} & 59.37 & 46.83 & 20.07 & 26.56 & 39.79 & 33.24 & 44.93 & \underline{43.64} \\
TADM\cite{long2026thermally}       & CVPR          & 2026 & 51.83 & 29.62 & \underline{62.74} & 25.13 & 7.55 & 16.56 & 26.64 & 11.84 & 39.31 & 30.14 \\
RFA-Tex\cite{zhu2026rfa}    & TIFS          & 2026 & 57.12 & 38.13 & 54.66 & \underline{49.61} & 13.77 & 27.99 & 32.16 & 17.17 & 23.89 & 34.94 \\
AdvReal\cite{huang2026advreal}    & ESWA          & 2026 & 31.96 & 23.98 & 28.17 & 30.18 & 6.22 & 18.85 & 20.16 & 5.31 & 9.74 & 19.40 \\
\midrule
\rowcolor{oursbg}
\textbf{MotionPatch} & -- & -- & \textbf{84.95} & \textbf{68.96} & \textbf{79.94} & \textbf{96.70} & \textbf{79.60} & \textbf{91.75} & \textbf{59.99} & \underline{34.22} & \textbf{59.85} & \textbf{72.88} \\
\bottomrule
\end{tabular}
\label{tab:whitebox-pedestrian}
\vspace{-1.5mm}

\end{table*}

\subsection{Experimental Setup}
\textbf{Dataset.}~We obtain driving scenes from the Waymo Open Dataset \citep{sun2020scalability} and reconstruct them with DriveStudio \citep{chen2025omnire}, each providing five calibrated camera views. The training split contains 70 scenes with 13,853 frames and the evaluation split contains 18 disjoint scenes with 3,572 frames. To evaluate generalization to unseen scenes and sensor configurations, we additionally use 10 scenes each from nuScenes \citep{caesar2020nuscenes} and Argoverse~2 \citep{Argoverse2}, reconstructed under the same protocol. Pedestrians are represented by 14 animatable avatars reconstructed from a single image that model pose-driven non-rigid clothing deformations. Seven avatars are pretrained models released with PERSONA \citep{sim2025persona}, the remaining seven are reconstructed from distinct HuGe100K subjects using IDOL \citep{zhuang2025idol}. Driven by SMPL-X \citep{pavlakos2019expressive} motion parameters retargeted from 38 AMASS \citep{mahmood2019amass} actions, these avatars are rendered within the reconstructed scenes along per-scene pedestrian trajectories to produce pedestrian observations across timestamps and camera views. All pedestrian trajectories are manually annotated and verified; details of the trajectory editing process are provided in the supplementary material.

\noindent\textbf{Implementation details.}~Training is conducted for 140 epochs on an NVIDIA RTX A6000 GPU. The patch resolution is $128\times128$. The Adam optimizer is used with learning rate 0.05, cosine annealing to $5\times10^{-4}$, and batch size 8. Training converges at epoch 140.


\noindent\textbf{Target detectors.}~We evaluate pedestrian detection attacks on COCO-trained YOLOv5\citep{jocher2020ultralytics}, YOLOv8\citep{yolov8_ultralytics}, YOLO26\citep{jocher2026ultralytics},
SSD\citep{liu2016ssd}, FCOS\cite{tian2019fcos}, RetinaNet\cite{lin2017focal},
Faster R-CNN\citep{ren2016faster}, Mask R-CNN\citep{he2017mask}, Cascade R-CNN\citep{cai2018cascade}, DETR\citep{carion2020end}, Deformable DETR (D-DETR)\citep{zhu2020deformable} and RT-DETR\citep{zhao2024detrs}, as well as language-conditioned models Grounding DINO(G-DINO)\citep{liu2024grounding}, LocateAnything-3B(LA-3B)\citep{wang2026locateanything}, Rex-Omni-3B(RO-3B)\citep{jiang2026detect} and VisionReasoner-7B(VR-7B)\citep{liu2026visionreasoner}. All models use publicly released checkpoints with frozen parameters, targeting the \emph{person} category or using person-oriented queries as appropriate.


\begin{table}[t]
\caption{Ablation study of MotionPatch components.}
\label{tab:ablation-study}
\centering
\small
\setlength{\tabcolsep}{3pt}
\begin{tabular}{ccccc}
\toprule
\textbf{3DGS} & \textbf{Motion} & \textbf{GAI} & \textbf{$\mathcal{L}_{\mathrm{cont}}$} & \textbf{ASR (\%) $\uparrow$} \\
\midrule
                & \checkmark & \checkmark & \checkmark & 62.09 \\
\checkmark      &            & \checkmark & \checkmark & 59.83 \\
\checkmark      & \checkmark &            & \checkmark & 71.24 \\
\checkmark      & \checkmark & \checkmark &            & 74.52 \\
\midrule
\rowcolor{gray!15}
\checkmark      & \checkmark & \checkmark & \checkmark & \textbf{84.95} \\
\bottomrule
\end{tabular}
\vspace{-3mm}
\end{table}

\noindent\textbf{Evaluation metrics.}~Following prior work~\cite{huang2026advreal}, we adopt five metrics to evaluate adversarial effectiveness: Attack Success Rate (ASR)~\citep{thys2019fooling}, Precision (P), Recall (R), AP@0.5, and F1-score. \textit{Further details are provided in the supplementary material.}

\begin{table*}[t]
\caption{Cross-model transferability (ASR \%). The adversarial patch is generated using YOLOv5}
\label{tab:cross-detector-transfer}
\centering
\resizebox{\textwidth}{!}{%
\setlength{\tabcolsep}{1.5pt}
\small
\begin{tabular}{l|ccccc@{\hspace{\tabcolsep}\vrule width0.4pt\hspace{\tabcolsep}}ccc@{\hspace{\tabcolsep}\vrule width0.4pt\hspace{\tabcolsep}}ccc@{\hspace{\tabcolsep}\vrule width0.4pt\hspace{\tabcolsep}}cccc}
\toprule
\multirow{2}{*}{\textbf{Method}}
& \multicolumn{5}{c}{\textbf{One-Stage}}
& \multicolumn{3}{c}{\textbf{Two-Stage}}
& \multicolumn{3}{c}{\textbf{Transformer-based}}
& \multicolumn{4}{c}{\textbf{VLM-based}} \\
\cmidrule(lr){2-6}
\cmidrule(lr){7-9}
\cmidrule(lr){10-12}
\cmidrule(lr){13-16}
& \textbf{YOLOv8}
& \textbf{YOLO26}
& \textbf{SSD}
& \textbf{FCOS}
& \textbf{RetinaNet}
& \textbf{M-RCNN}
& \textbf{F-RCNN}
& \textbf{C-RCNN}
& \textbf{DETR}
& \textbf{D-DETR}
& \textbf{RT-DETR}
& \textbf{G-DINO}
& \textbf{LA-3B}
& \textbf{RO-3B}
& \textbf{VR-7B} \\
\midrule
AdvPatch
& 35.97 & 46.29 & 46.69 & 35.71 & 42.40
& 26.73 & 28.52 & 46.17
& 25.56 & 40.16 & 9.67
& 11.96 & 9.06 & 8.38 & 11.08 \\
NatPatch
& 47.30 & 43.35 & 45.59 & 54.53 & 58.78
& 36.03 & 39.01 & 33.30
& 39.58 & 54.32 & 11.90
& 21.21 & 14.63 & 11.52 & 10.99 \\
AdvTexture
& 34.32 & 40.70 & 34.52 & 25.59 & 34.57
& 15.36 & 17.96 & 62.79
& 20.68 & 26.83 & 7.50
& 15.02 & 5.49 & 7.68 & 7.69 \\
T-SEA
& 49.85 & 47.36 & 45.30 & 53.64 & 57.07
& 39.28 & 41.88 & 38.00
& 39.08 & 55.51 & 13.47
& 19.17 & 13.72 & 12.74 & 10.53 \\
DM
& 51.88 & 50.74 & 48.08 & 56.83 & 59.30
& 39.29 & 41.07 & 65.86
& 35.89 & 53.37 & 11.60
& 21.08 & 10.89 & 12.30 & 11.26 \\
AdvReal
& 26.19 & 30.88 & 21.80 & 17.13 & 24.16
& 9.32 & 10.26 & 17.46
& 13.19 & 18.14 & 4.88
& 8.48 & 1.91 & 3.23 & 7.23 \\
\midrule
\rowcolor{oursbg}
\textbf{MotionPatch}
& \textbf{79.28} & \textbf{80.91} & \textbf{82.46} & \textbf{75.92} & \textbf{78.10}
& \textbf{65.18} & \textbf{68.96} & \textbf{72.64}
& \textbf{63.92} & \textbf{79.94} & \textbf{36.47}
& \textbf{39.76} & \textbf{28.68} & \textbf{29.32} & \textbf{19.05} \\
\bottomrule
\end{tabular}%
}
\vspace{-3mm}
\end{table*}

\begin{figure}[t]
    \centering
    \includegraphics[width=\linewidth]{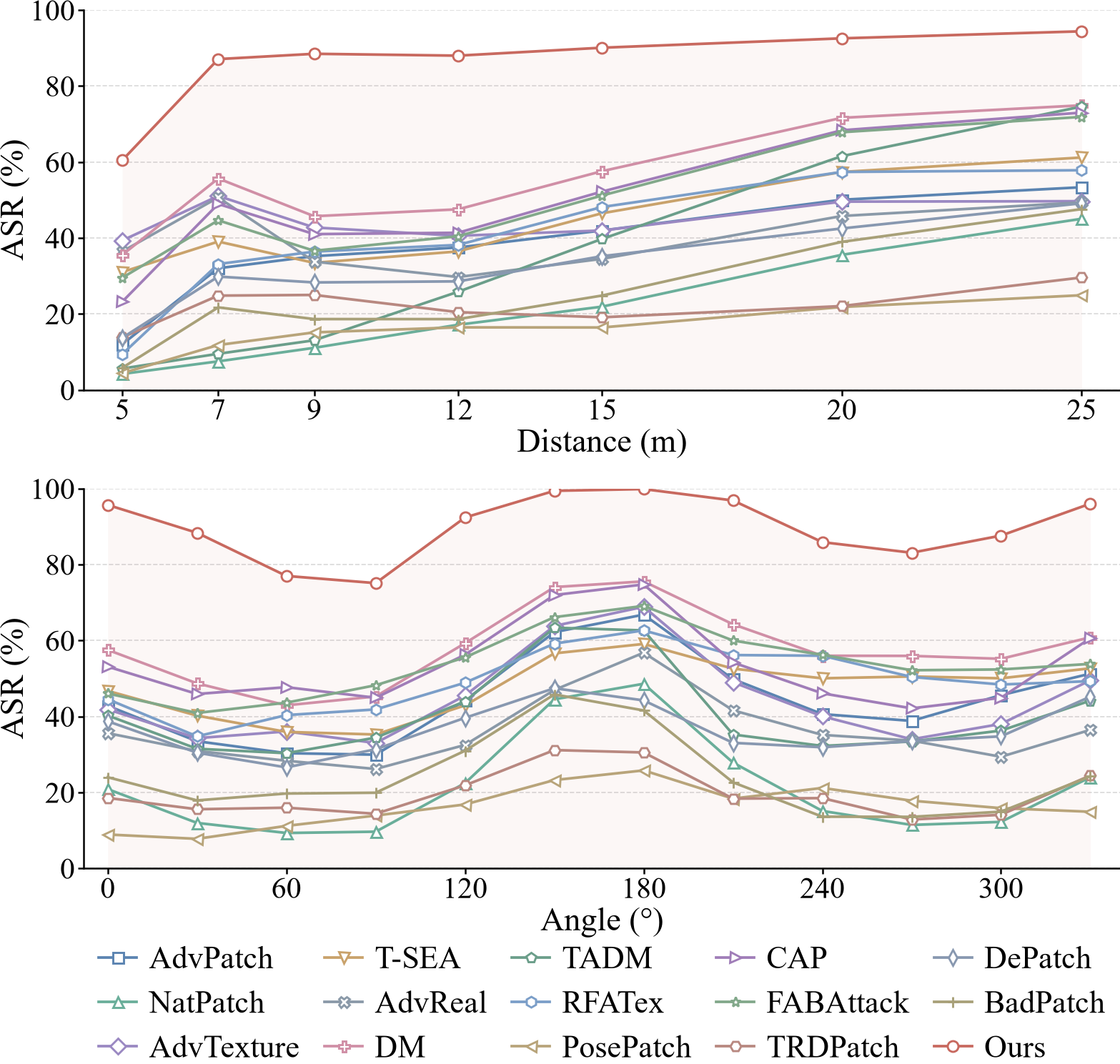}
    \caption{Comparison of detection results for different attack methods across various distances and azimuth angles, specifically reporting the average ASR over diverse scenes.}
    \label{fig:placeholder}
    \vspace{-5mm}
\end{figure}

\subsection{Digital experiments}
Table~\ref{tab:whitebox-pedestrian} compares MotionPatch against baseline across three datasets and three detectors under the white-box setting. MotionPatch achieves an average ASR of 72.88\%, outperforming the strongest baseline, FABAttack, by 29.24 points. The margin is largest on nuScenes, where MotionPatch reaches 96.70\% on YOLOv5 while the strongest competing method, RFA-Tex, reaches 49.61\%. On Waymo, the margin ranges from 10 to 18 points across detectors, and on Argoverse~2 MotionPatch achieves the highest ASR in most comparisons. AdvReal, despite using 3D rendering, achieves only 16.75\% on average, confirming that 3D rendering alone is insufficient without scene-consistent optimization.

Figure~\ref{fig:placeholder} reports ASR as a function of capture distance and viewing angle in digital simulation. MotionPatch reaches near 100\% ASR at 25\,m while baselines plateau below 70\%, and it sustains high ASR across all viewing angles with only a modest dip at side-on views. Baselines show sharper drops at the same angles, indicating stronger dependence on frontal appearance.

\noindent\textbf{Ablation study.}~Table~\ref{tab:ablation-study} removes one component at a time from the full model. Removing motion causes the largest drop of 25.12 points, since a patch optimized under a single pose does not account for the limb and garment deformation that occurs during walking. Removing the reconstructed 3DGS scene reduces ASR by 22.86 points, as the patch is no longer optimized under depth-consistent occlusion and real scene context. Removing GAI reduces ASR by 13.71 points, because the patch is optimized under uniform lighting rather than the local illumination of each view. Removing $\mathcal{L}_{\mathrm{cont}}$ reduces ASR by 10.43 points, as the optimization no longer targets frames on which the detector briefly recovers. The advantage of MotionPatch arises from the combination of these components rather than any single one.

\begin{table}[t]
\caption{Attack success rate (ASR, \%) across five pedestrian motion sequences: (A) walking, (B) running, (C) walk--hop--walk, (D) walk--run transitions, and (E) walk--leap--walk.}
\label{tab:asr-action-compare}
\centering
\setlength{\tabcolsep}{4pt}
\small
\renewcommand{\arraystretch}{1}
\begin{tabular}{l|ccccc|c}
\toprule
\textbf{Method} & \textbf{A} & \textbf{B} & \textbf{C} & \textbf{D} & \textbf{E} & \textbf{Avg.} \\
\midrule
AdvPatch   & 35.39 & 37.96 & 35.85 & 48.04 & 40.54 & 39.56 \\
NatPatch   & 49.17 & 50.51 & 49.26 & 60.97 & 51.90 & 52.36 \\
AdvTexture & 48.13 & 50.19 & 49.94 & 60.43 & 49.42 & 51.62 \\
T-SEA      & 51.96 & 54.86 & 52.19 & 62.13 & 55.25 & 55.28 \\
DePatch    & 37.45 & 38.94 & 35.24 & 45.83 & 41.99 & 39.89 \\
CAP        & 59.64 & 61.01 & 62.60 & 75.85 & 65.63 & 64.95 \\
DM         & 52.90 & 54.04 & 53.83 & 61.15 & 52.94 & 54.97 \\
BadPatch   & 33.84 & 34.03 & 34.24 & 45.86 & 37.40 & 37.07 \\
PosePatch  & 25.48 & 26.09 & 21.92 & 32.09 & 29.38 & 26.99 \\
TRDPatch   & 28.42 & 28.09 & 27.10 & 41.94 & 33.88 & 31.89 \\
FABAttack  & 55.04 & 54.59 & 54.25 & 66.25 & 62.75 & 58.58 \\
TADM       & 46.65 & 49.21 & 44.81 & 60.01 & 52.16 & 50.57 \\
RFA-Tex    & 49.44 & 49.37 & 49.47 & 61.31 & 55.49 & 53.02 \\
AdvReal    & 32.01 & 32.31 & 35.12 & 40.14 & 34.86 & 34.89 \\
\midrule
\rowcolor{oursbg}
\textbf{MotionPatch} & \textbf{77.25} & \textbf{79.10} & \textbf{78.96} & \textbf{88.97} & \textbf{77.33} & \textbf{80.32} \\
\bottomrule
\end{tabular}
\vspace{-3mm}
\end{table}

\begin{figure*}
    \centering
    \includegraphics[width=\linewidth]{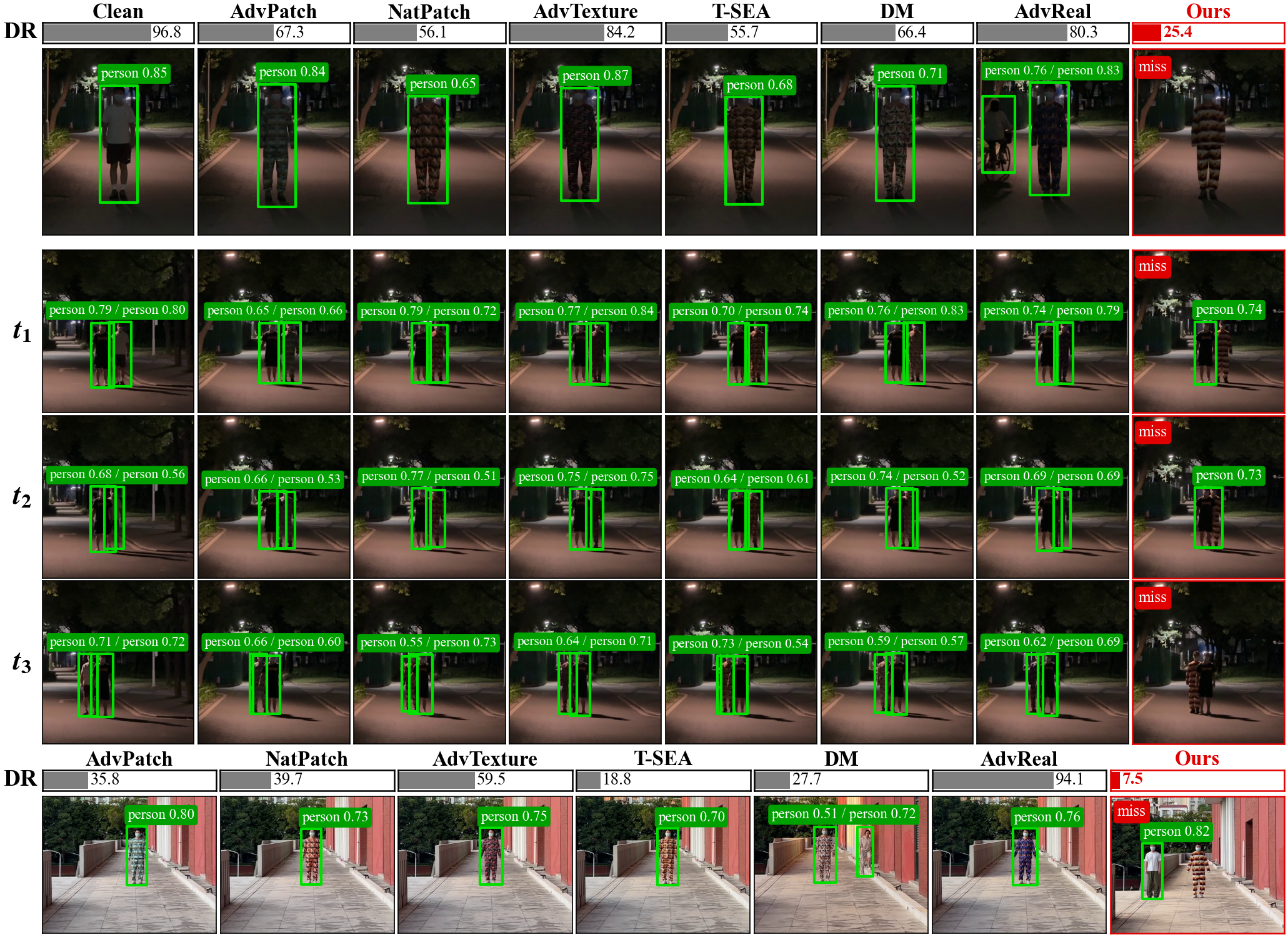}
    \caption{Physical attack on a walking pedestrian under varying conditions. The first row shows results at close range under extremely low brightness. Rows 2--4 show results under normal pedestrian walking with different occlusion ratios. The bottom row shows results under daytime conditions. Detection Rate (DR, \%) over the full video is reported below each method.}
    \label{fig:physical-night}
    \vspace{-3mm}
\end{figure*} 

\noindent\textbf{Transferability.}~Table~\ref{tab:cross-detector-transfer} evaluates patches optimized on YOLOv5s against 15 black-box models from four architecture families. MotionPatch achieves the highest ASR on all 15 detections. On one-stage detectors, which share the dense-prediction design of the source, MotionPatch leads the strongest baseline by 18.80 to 34.38 points. Because the patch is optimized across varying viewpoints, poses, occlusions, and illumination, it depends less on the specific responses of YOLOv5, which helps it remain robustness on unseen architectures.

With the rapid progress of vision-language models, we further test whether the patch transfers to four VLM-based detectors: G-DINO, LA-3B, RO-3B, and VR-7B. These models localize pedestrians from text queries rather than fixed-category heads, and three of them contain 3B to 7B parameters, so they differ from YOLOv5s in both scale and architecture. Although all methods lose effectiveness on these models, MotionPatch still leads every baseline by a large margin, reaching an average ASR of 29.20\% compared with 14.59\% for the best baseline, NatPatch.  This indicates that the advantage of MotionPatch becomes more pronounced as the target moves further from the source architecture.

\noindent\textbf{Robustness across pedestrian motions.}~Table~\ref{tab:asr-action-compare} evaluates five motion sequences, from steady walking and running to transitions with abrupt pose changes such as hops and leaps. MotionPatch achieves the highest ASR on every sequence, with an average of 80.32\% compared with 64.95\% for the strongest baseline, CAP. MotionPatch is also the most stable across motion types. All methods reach their highest ASR on the walk--run transition (D), and across the remaining four sequences MotionPatch varies by only 1.85 points, while CAP and FABAttack vary by 5.99 and 8.50 points. In particular, the hop (C) and leap (E) sequences, which introduce sudden changes in limb configuration and body height, yield 78.96\% and 77.33\%, on par with steady walking. 



\subsection{Physical Experiments}

Figure~\ref{fig:physical-night} shows physical attack results on a walking pedestrian under varying lighting and occlusion conditions. Green bounding boxes indicate successful detections by YOLOv5s, while red labels mark suppressed detections. Each method's detection rate over the full sequence is reported below its column. Under night conditions, MotionPatch achieves a DR of 25.4\%, the lowest among all compared methods and 30.3 points below the second best, T-SEA at 55.7\%. In the middle rows, where the pedestrian with different occlusion ratios, MotionPatch consistently suppresses detections while most baselines maintain multiple detections per frame. Under daytime conditions, MotionPatch achieves a DR of 7.5\%, while every baseline exceeds 20\%. This consistent suppression across lighting and occlusion conditions can be attributed in part to two design choices: 3D Geometry-Aware Illumination adapts the patch to varying scene lighting during optimization, and training under diverse SMPL-X motion sequences accounts for limb configurations and garment deformations. AdvReal, despite using 3D rendering, reaches a DR of 80.3\% under night conditions and 94.1\% under daytime conditions, suggesting that scene-integrated optimization matters beyond 3D rendering alone. \textit{Additional physical experiments are provided in the supplementary material and videos.}
\section{Conclusion}

In this paper, we presented MotionPatch, a motion-aware adversarial patch optimization framework that addresses key limitations of existing physical attacks on person detectors. First, we model pedestrian motion as continuous trajectories within reconstructed 4D driving scenes, capturing motion-induced changes in pose, viewpoint, and occlusion during optimization. Second, we enforce depth-consistent occlusion between inserted pedestrians and scene geometry through differentiable Gaussian splatting, and introduce 3D Geometry-Aware Illumination to match pedestrian appearance to local scene lighting. Third, we propose a temporal continuity loss that couples consecutive observations of the same pedestrian, suppressing brief detection recoveries that aggregate metrics may overlook. Experiments demonstrate that MotionPatch achieves 84.95\% white-box ASR on YOLOv5s and maintains transferability across conventional detectors, vision-language models, and diverse motion sequences. Physical experiments show its effectiveness under varying lighting and occlusion conditions. These results demonstrate that scene-grounded, motion-aware optimization provides a more realistic evaluation of adversarial vulnerabilities in autonomous driving perception systems.
\section*{LLM Usage Statement}
We used DeepSeek to assist with drafting, restructuring, and
language editing of the manuscript text, including related-work polishing. LLMs were not used to generate research ideas, design experiments, create or annotate data, or
run experiments. All LLM-assisted content was reviewed and verified by the
authors, who take full responsibility for all submitted content.
{
    \small
    \bibliographystyle{ieeenat_fullname}
    \bibliography{main}
}


\end{document}